# Study of a committee of neural networks for biometric hand-geometry recognition

Marcos Faundez-Zanuy

Escola Universitària Politècnica de Mataró
Universitat Politècnica de Catalunya, BARCELONA (SPAIN)
`faundez@eupmt.es`
`http://www.eupmt.es/veu`

**Abstract**. This Paper studies different committees of neural networks for biometric pattern recognition. We use the neural nets as classifiers for identification and verification purposes. We show that a committee of nets can improve the recognition rates when compared with a multi-start initialization algorithm that just picks up the neural net which offers the best performance. On the other hand, we found that there is no strong correlation between identification and verification applications using the same classifier.

## 1  Introduction

Neural networks are one of the most powerful tools for pattern recognition [1]. On the other hand, one of the applications of pattern recognition, person identification by means of biometrics [2], has gain market share. Nevertheless, still remain some unsolved problems [3-5].

In this paper, we focus on a biometric identification system improvement by means of a committee of neural networks. Figure 1 shows the general scheme of a biometric system.

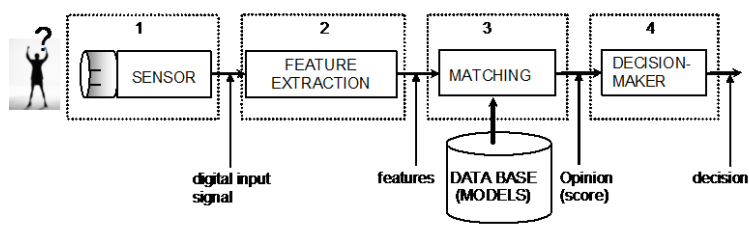

**Fig. 1.** General scheme of a biometric recognition system.

These systems can be operated in two ways:
a) Identification: In this approach no identity is claimed from the person. The automatic system must determine who is trying to access.
b) Verification: In this approach the goal of the system is to determine whether the person is who he/she claims to be. This implies that the user must provide an identity and the system just accepts or rejects the users according to a successful or unsuccessful verification. Sometimes this operation mode is named authentication or detection.

We will focus on a study of the third block of figure 1. Without loss of generality we will use a set of features and a database extracted from hand-geometry images obtained from a 22 people and 10 different realizations per person set (5 for training and 5 for testing). The feature extraction section and the digital signal input (blocks 1 and 2 of figure 1) can be found in [6]. Our matching algorithm will be a neural network or a committee of neural networks.

### 1.1 Biometric identification

The system performance can be evaluated using an identification rate. This is obtained by the ratio between the number of properly assigned identities and the number of trials of the experiment. For this purpose, having a labelled training set that contains examples of each person, we train a neural network in a discriminative mode: for each input pattern we force the neural net to learn +1 at the output of the corresponding user and –1 at the other outputs.

$$O[i] = \begin{cases} 1 & if \ \vec{x} = (x[1], x[2], \cdots, x[P]) \in person\#i \\ -1 & otherwise \end{cases} \quad (1)$$

Figure 2 shows the neural network scheme.

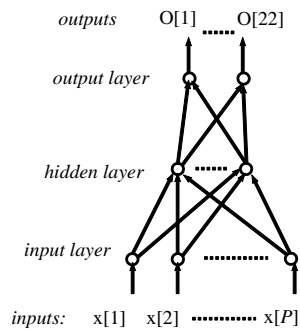

**Fig. 2.** Neural network architecture.

For identification, if we have a population of $N$ different people, and a labelled test set, we fill a matrix S, where the elements are interpreted in the following way:

$$s_{ijk} = O[j]\big|_{\vec{x} \in person\#i} \quad, k = 1, \cdots \#trials \quad (2)$$

Where trials is the number of different testing images per person ($k$=5 in our experiments), and $s_{ijk}$ is the similarity from the $k$ realization of an input signal belonging to person $i$, to the model of person $j$.

This matrix can be drawn as a three dimensional data structure (see figure 3).

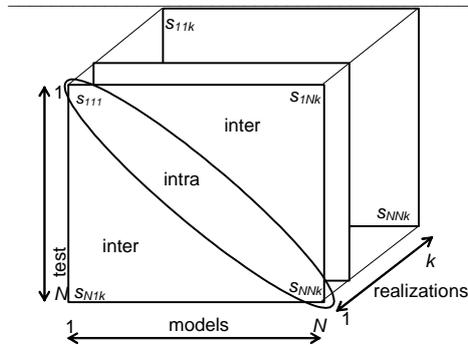

**Fig. 3.** Proposed data structure.

Thus, the identification rate looks for each realization, in each raw, if the maximum similarity is inside the principal diagonal (success) or not (error), and works out the identification rate as the ration between successes and number of trials (successes + errors):

```
for i=1:N,
  for k=1:#trials,
    if(s_iik>s_ijk)  ∀j≠i, then success=success+1
    else error=error+1
    end
  end
end
```

### 1.2 Biometric verification

Verification systems can be evaluated using the False Acceptance Rate (FAR, those situations where an impostor is accepted) and the False Rejection Rate (FRR, those situations where a user is incorrectly rejected), also known in detection theory as False Alarm and Miss, respectively. This framework gives us the possibility of distinguishing between the discriminability of the system and the decision bias. The discriminability is inherent to the classification system used and the discrimination bias is related to the preferences/necessities of the user in relation to the relative importance of each of the two possible mistakes (misses vs. false alarms) that can be done in verification. This trade-off between both errors has to be usually established by adjusting a decision threshold. The performance can be plotted in a ROC (Receiver Operator Characteristic) or in a DET (Detection error trade-off) plot [7]. DET curve gives uniform treatment to both types of error, and uses a scale for both axes, which spreads out the plot and better distinguishes different well performing systems and usually produces plots that are close to linear. DET plot uses a logarithmic scale that expands the extreme parts of the curve, which are the parts that give the most information about the system performance. For this reason the speech community prefers DET instead of ROC plots. Figure 4 shows an example of DET plot.

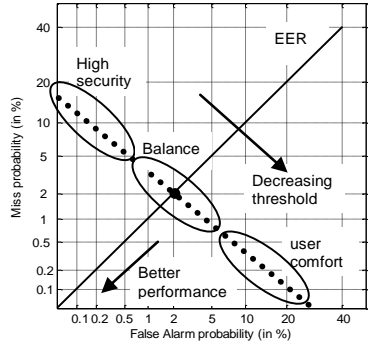

**Fig. 4.** Example of a DET plot for a user verification system (dotted line).

We have used the minimum value of the Detection Cost Function (DCF) for comparison purposes. This parameter is defined as [7]:

$$DCF = C_{miss} \times P_{miss} \times P_{true} + C_{fa} \times P_{fa} \times P_{false} \qquad (3)$$

Where $C_{miss}$ is the cost of a miss (rejection), $C_{fa}$ is the cost of a false alarm (acceptance), $P_{true}$ is the a priori probability of the target, and $P_{false} = 1 - P_{true}$. $C_{miss} = C_{fa} = 1$.

Using the data structure defined in figure 3, we can easily apply the DET curve analysis. We just need to split the distances into two sets: intra-distances (those inside the principal diagonal), and inter-distances (those outside the principal diagonal).

## 2   Experimental results and conclusions.

We have trained a neural network on the conditions of previous section, and evaluated the identification rates and minimum value of the DCF. The experiments have been done with the following parameters:
   a)   Multi-Layer Perceptron 9×30×22 trained with the Levengerb-Marquardt algorithm during 10 epochs.
   b)   Multi-Layer Perceptron 9×30×22 trained with the Levengerb-Marquardt algorithm during 50 epochs using regularization [8].
   c)   A committee of three MLP, each one trained on the conditions of section a)
   d)   A committee of three MLP, each one trained on the condition of section b)

### 2.1   Committee of neural networks

In pattern recognition applications it is well known that a number of differently trained neural networks (that can be considered as "experts"), which share a common input, can produce a better result if their outputs are combined to produce an overall output. This technique is known as ensemble averaging, committee machine [9], data fusion [10], etc. The motivation for its use is twofold [9]:

- If the combination of experts were replaced by a single neural network, the number of equivalent adjustable parameters would be large, and this implies more training time and local minima problems [11].
- The risks of over-fitting the data increases when the number of adjustable parameters is large compared to the size of the training data set.

Figure 5 shows the equivalent scheme for an *M* experts' combination. We have considered each classifier as an expert with scalar output $y_i[n]$ and vectorial input $\vec{x}$:

$$y_i[n] = F_i(\vec{x}) = F_i(x[n-1], x[n-2], \cdots, x[n-P]) \tag{4}$$

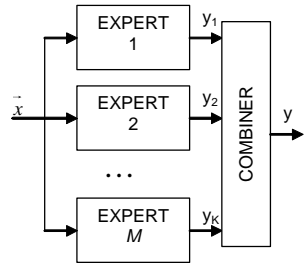

**Fig. 5.** Committee machine based on ensemble-averaging

The expectation is that differently trained experts converge to different local minima on the error surface, and overall performance is improved by combining the outputs in some way. Let $\vec{x}$ denote an input vector and $y_d$ the desired response (predicted output value) for this input value. If the output of the neural net number *i* is $F_i(\vec{x})$, we can compute the mean squared error (MSE) of the obtained output as:

$$MSE\left[F_i(\vec{x})\right] = E\left[\left(y_d - F_i(\vec{x})\right)^2\right] \tag{5}$$

We can average the performances for a set of different neural networks (or the same architecture trained with the same algorithm but using a different random initialization) to obtain the average mean square error:

$$E\left[MSE\left[F_i(\vec{x})\right]\right] = \frac{1}{N}\sum_{i=1}^{M} E\left[\left(y_d - F_i(\vec{x})\right)^2\right] \tag{6}$$

We can define the Basic Ensemble Method (BEM) regression function [12] as:

$$F_{BEM}(\vec{x}) \equiv \frac{1}{M}\sum_{i=1}^{M} F_i(\vec{x}) = y_d - \frac{1}{M}\sum_{i=1}^{M}\left(y_d - F_i(\vec{x})\right) \tag{7}$$

If we assume that the errors $m_i = y_d - F_i(\vec{x})$, $i=1,\ldots M$ of each individual predictor are mutually independent with zero mean, we can calculate the MSE of $F_{BEM}(\vec{x})$:

$$MSE\left[F_{BEM}(\vec{x})\right] = E\left[\left(\frac{1}{M}\sum_{i=1}^{M} m_i\right)^2\right] = \frac{1}{M^2} E\left[\sum_{i=1}^{M} m_i^2\right] + \frac{1}{M^2} E\left[\sum_{\substack{i=1 \\ i \neq j}}^{M} m_i m_j\right] = \quad (8)$$

$$\frac{1}{M^2} E\left[\sum_{i=1}^{M} m_i^2\right] + \frac{1}{M^2} \sum_{\substack{i=1 \\ i \neq j}}^{M} E[m_i] E[m_j] = \frac{1}{M^2} E\left[\sum_{i=1}^{M} m_i^2\right] \Rightarrow MSE[F_{BEM}] = \frac{1}{M} E[MSE] \quad (9)$$

This result shows that, by averaging regression estimates, we can reduce the MSE by a factor of *M* when compared to the individual population performance. By increasing the population size, we can, in principle, make the estimation error arbitrarily small, but if $M \to \infty$ the assumption that the errors are mutually independent is not valid. In practice there is saturation on performance, and after a given value of *M* there is no improvement with an increase on the number of combined neural net classifiers. In our experiments, we will use *M*=3.

### 2.2 Results and conclusions.

We have done 100 random initializations for each one of the four schemes defined in the introductory part of section 2. Figure 6 shows the histograms for Identification rate and minimum value of DCF (left and right respectively). Figure 7 shows a fitted Gaussian to the previous histograms (fig. 6). It is clear that the committee of experts improves the mean value and decreases the variance. Thus, with a committee of experts a better recognition result is achieved.

Another interesting study is the correlation between identification and verification rates. That is: if we get a good system for identification, does it mean that we have a good verifier? Table 1 shows the identification and verification results for each studied scenario along with the correlation factor between both parameters.

Although these results are not as good as other biometric traits such us fingerprints [13-16], for our purpose we prefer this set of experimental data, where the classifier produces more errors. Figure 8 shows a two-dimensional plot, where for each classifier, the identification versus the verification results are plotted. Surprisingly, we observe that the best classifier for identification does not necessarily imply the best performance for verification. We have also interpolated a first-order polynomial in order to see the tendency. Fortunately, the best the classifier for identification is, the better the tendency for verification. However, there are several classifiers that achieve the best identification rate, and the equivalent DCF can differ significantly. We think that an interesting research line should solve the question: what needs a good identifier in order to also be a good verifier? This question is not easy to solve. However, we have observed that regularization provides higher correlation (absolute value of the correlation coefficient) between a good classifier and a good verifier than its counterpart without regularization.

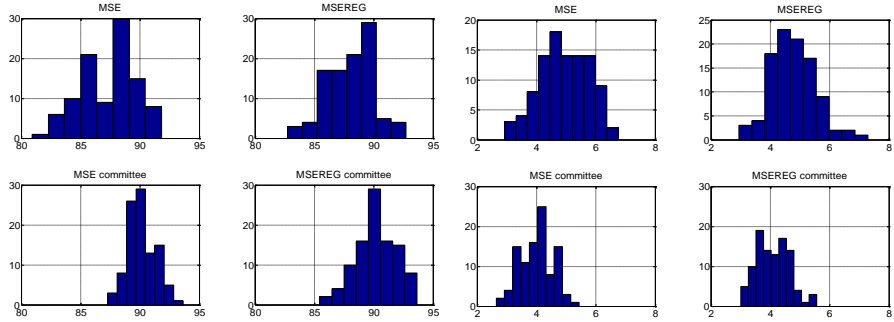

**Fig. 6.** histograms of identification rates (on the left) and DCF (on the right) for different classification strategies and 100 random initializations.

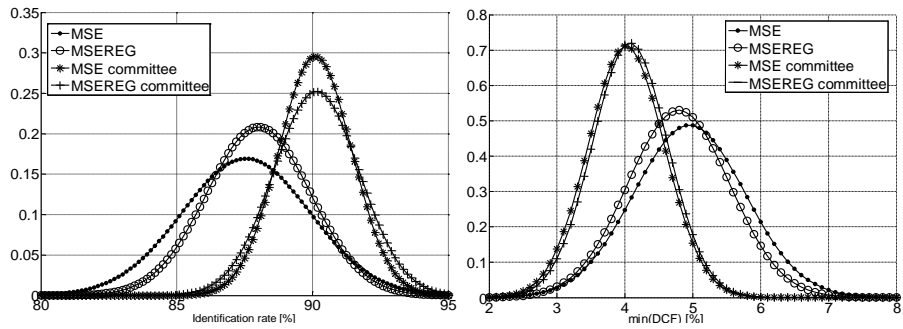

**Fig. 7.** A fitted Gaussian to each histogram of figure 6.

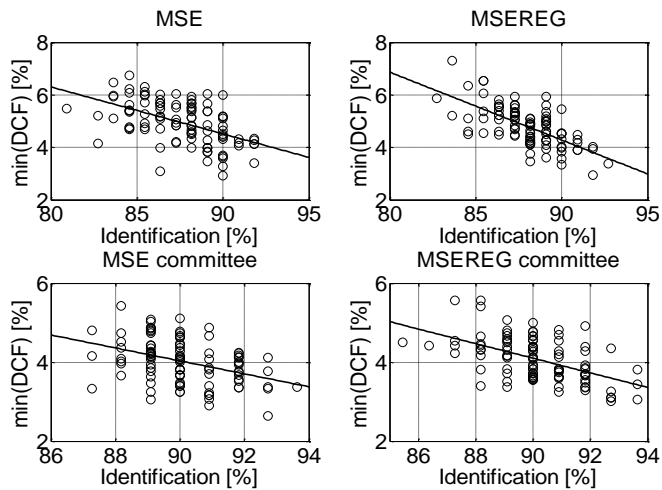

**Fig. 8.** 2-D plot of identification versus DCF pair values for each trained classifier.

**Table 1.** Comparison between different schemes performance

| Classifier | Identification rate (%) | Min(DCF) | Corr(Ident,DCF) |
|---|---|---|---|
| MLP (MSE) | 91,82% | 2.92% | –0.52 |
| MLP (MSEREG) | 92,73% | 2.94% | –0.66 |
| MLP (MSE) committee | 93,64% | 2.64% | –0.39 |
| MLP (MSEREG) committee | 93,64% | 3.01% | –0.53 |

## Acknowledgement

This work has been supported by FEDER and the Spanish grant MCYT TIC2003-08382-C05-02.